\def\year{2020}\relax
\documentclass[letterpaper]{article} 
\usepackage{aaai20}  
\usepackage{times}  
\usepackage{helvet} 
\usepackage{courier}  
\usepackage[hyphens]{url}  
\usepackage{graphicx} 
\usepackage{comment}
\usepackage{booktabs}
\usepackage{amsfonts}
\usepackage{subcaption}
\usepackage{adjustbox}      
\usepackage{tabulary}
\usepackage{multirow}
\usepackage{mdwlist}

\usepackage{graphicx}  
\title{Interpretable Rumor Detection in Microblogs by Attending to User Interactions}
\author{Ling Min Serena Khoo, Hai Leong Chieu \\ DSO National Laboratories \\ 12 Science Park Drive \\ Singapore 118225 \\ \{klingmin,chaileon\}@dso.org.sg \And Zhong Qian, Jing Jiang \\ Singapore Management University \\ 80 Stamford Road \\ Singapore 178902 \\ qianzhongqz@163.com,jingjiang@smu.edu.sg}

\begin{document}

\maketitle

\begin{abstract}
We address rumor detection by learning to differentiate between the community's response to real and fake claims in microblogs. Existing state-of-the-art models are based on tree models that model conversational trees.
However, in social media, a user posting a reply might be replying to the entire thread rather than to a specific user. We propose a post-level attention model (PLAN) to model long distance interactions between tweets with the multi-head attention mechanism in a transformer network. 
We investigated variants of this model: (1) a structure aware self-attention model (StA-PLAN) that incorporates tree structure information in the transformer network, and (2) a hierarchical token and post-level attention model (StA-HiTPLAN) that learns a sentence representation with token-level self-attention. To the best of our knowledge, we are the first to evaluate our models on two rumor detection data sets: the PHEME data set as well as the Twitter15 and Twitter16 data sets. We show that our best models outperform current state-of-the-art models for both data sets. Moreover, the attention mechanism allows us to explain rumor detection predictions at both token-level and post-level.
\end{abstract}




\newcommand{\unverified}{{\it unverified}}
\newcommand{\nonrumor}{{\it non-rumor}}
\newcommand{\falserumor}{{\it false-rumor}}
\newcommand{\truerumor}{{\it true-rumor}}

\newcommand{\citet}[1]{\citeauthor{#1}~\shortcite{#1}}
\newcommand{\citep}{\cite}
\newcommand{\citealp}[1]{\citeauthor{#1}~\citeyear{#1}}

\maketitle

\section{Introduction}

The spread of fake news can have far reaching and devastating effects. \$130 billion in stock value was wiped out in minutes after a false Associated Press's tweet claimed that Barack Obama was injured following an explosion in 2013 ~\citep{rapoza_2017}. Some researchers even believe that fake news has affected the outcome of the 2016 United States presidential election \cite{gunther2018fake}. The severity of the impact of fake news warrants the need for an effective and automated means of detecting fake news and has hence spurred much research in this area.

\begin{figure}[t]
\centering
    \includegraphics[width= 0.98\columnwidth]{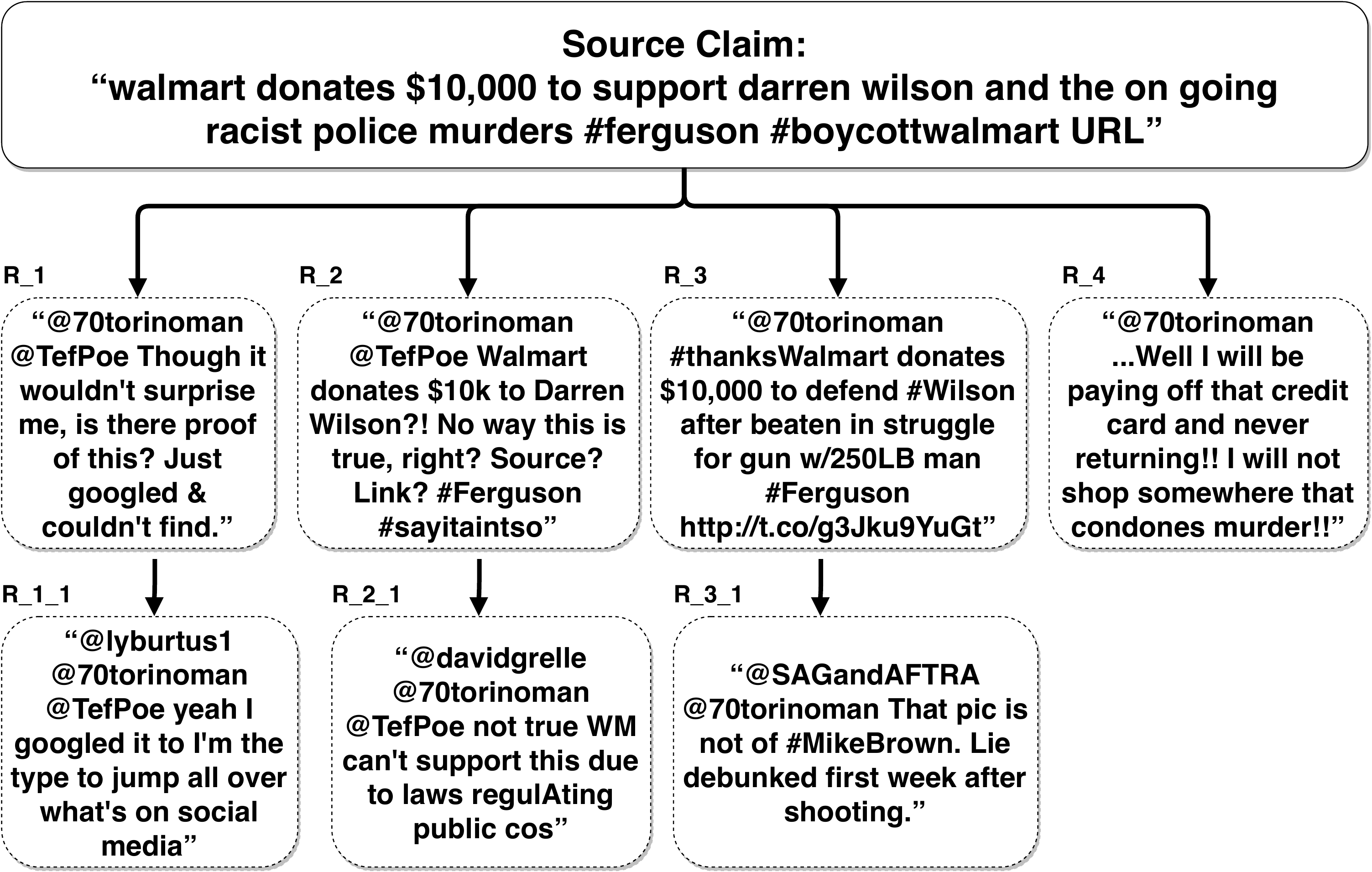}
\caption{Sample of a thread from Twitter15 resulting from a fake claim.}
\label{fig:proptree}
\end{figure}

Our work focuses on the detection of fake claims using community response to such claims. This area of research exploits the collective wisdom of the community by applying natural language processing to comments directed towards a claim (see Figure~\ref{fig:proptree}). The key principle behind such works is that users on social media would share opinions, conjectures and evidences for inaccurate information. Hence, the interaction between users as well as the content shared could be captured for fake news detection. We discuss briefly two state-of-the-art models ~\citep{ma18,kumar19}.

Ma et al.~\shortcite{ma18} organized the source claim and its responding tweets in a tree structure as shown in Figure~\ref{fig:proptree}. Each tweet is represented as a node; the top node would be the source claim and the children of a node are tweets that have responded to it directly. They modeled the spread of information in a propagation tree using recursive neural networks. Signals from different nodes are aggregated recursively in either a bottom-up or a top-down manner. Information is propagated from the child node to the parent node in the bottom-up model and vice versa in a top-down model.
Similarly, Kumar et al.~\shortcite{kumar19} organized conversation threads with a tree structure and explored several variants of branch and tree LSTMs for rumour detection. 

Both papers used tree models with the intention of modelling structural information present in the conversation thread. Information is propagated either from the parent to the child or vice versa in tree models. However, the thread structure in social media conversations might be unclear. Each user is often able to observe all the replies in different branches of the conversation. A user debunking a fake news may not be directed solely at the person he is replying to - the content created could also be applicable to other tweets in the thread. Tree models do not model interactions between nodes from other branches explicitly and is a key limitation when modelling social media conversations. For example, in Figure \ref{fig:proptree}, tweet \textit{R\_1} and its replying tweet, \textit{R\_1\_1}, have expressed doubt regarding the factuality of the source claim. Tweets \textit{R\_2\_1} and \textit{R\_3\_1} have provided conclusive evidence that debunk the source claims as fake. Though \textit{R\_2\_1} and \textit{R\_3\_1} are child nodes of \textit{R\_2} and \textit{R\_3} respectively, they could provide important information to all the other nodes along the tree, such as \textit{R\_1\_1} and \textit{R\_1}. Therefore, we should consider interactions between all tweets, not just those between parent and their child, for better aggregation of information from the replying tweets. 

In this paper, we propose to overcome some limitations of tree models in modelling social media conversations as identified from our analysis. More specifically, we flattened the tree structure and arranged all tweets in a chronological order. We propose a post-level attention model (PLAN) that allows all possible pairwise interaction between tweets with the self-attention mechanism. To combine the strengths of both self-attention mechanism and tree models, we incorporated structural information and performed structure aware self-attention at a post-level (StA-PLAN). Lastly, we designed a structure aware hierarchical token and and post-level attention network (StA-HiTPLAN) to learn more complex sentence representations for each tweet. \newline
\newline 
The contributions of this paper are the following:
\begin{itemize}
    \item We utilize the attention weights from our model to provide both token-level and post-level explanations behind the model's prediction. To the best of our knowledge, we are the first paper that has done this.
    \item We compare against previous works on two data sets - PHEME 5 events \cite{kochkina_liakata_zubiaga_2018} and Twitter15 and Twitter16 \cite{ma17}. Previous works only evaluated on one of the two data sets. 
    \item  Our proposed models could outperform current state-of-the-art models for both data sets.
\end{itemize}

The rest of the paper is organized as follows. In Section 2, we examine some related work. We define our problem statement in Section 3. We present our models in Section 4 and results in Section 5. We then conclude with future work in Section 6.

\section{Related Work}

Existing approaches on automatically differentiating fake from real claims leverage on a variety of features: (i) the  content of the claim, (ii) the bias and social network of the source of the claim,  (iii) fact checking with trustworthy sources (e.g., Wikipedia), and (iv) community response to the claims. Our work falls into the last class of detecting rumors or fake claims from community response. In this section, we give a brief review of each class of work, focusing on works that detect fake claims using community response. For a more detailed survey, we refer the reader to~\cite{sharma2019}.

\textbf{Content Information}: Early work on deceptive content detection studied the use of linguistic cues such as percentage of pronouns, word length, verb quantity, and word classes~\cite{fuller09,mihalcea09,ott11,rubin16} on fake reviews, witness accounts, and satire. The detection of fake news using linguistic features have also been studied in \cite{obrien18,wang17liar}. Such analysis of deceptive content relies on linguistic features which might be unique to domains or topics. 

\textbf{Source and Social Network}:
Another group of work studied the source of fake news, and its social network. Wang et al.~\shortcite{wang17liar} found that adding source information to the content improves fake news classification accuracy. Hu et al.~\shortcite{hu13} found that accounts created to spread fake news tend to have different social network characteristics.

\textbf{Fact Checking}:
Fact checking websites such as \url{politifact.com} and \url{snopes.com} rely on manual verification to debunk fake news, but are unable to match up the rate at which fake news are being generated \cite{poynter_2019}. Automated fact checking aims to check claims against trustworthy sources such as Wikipedia~\cite{giovanni15}.  More recently, Thorne et al.~\shortcite{thorne18} proposed the FEVER shared task to verify an input claim against a database of 5 million Wikipedia documents, and classify each claim into one of three classes: \textit{Supports}, \textit{Refute} or \textit{Not Enough Info}. Fact checking is a more principled approach to fake news detection. However, it also requires an established corpus of verified facts and may not work for new claims with little evidence. 

\textbf{Community Response}:
Detecting fake news from community response is most closely related to our paper. Researchers have worked on automatically predicting the veracity of a claim by building classifiers that leverages on the comments and replies to social media posts, as well as the propagation pattern~\cite{omar17,castillo11,ma17,ma18}. Ma et al.~\shortcite{Ma:2018:DRS:3184558.3188729} adopted a multi-task learning approach to build a classifier that learns stance aware features for rumour detection. Similarly, Li et al.~\shortcite{li-etal-2019-rumor} adopted a multi-task learning approach for his models and included user information in his models. Chen et al. ~\shortcite{DBLP:journals/corr/ChenWLZYW17} proposed to pool out distinctive features that captures contextual variations of posts over time. Beyond linguistic features, other researchers have also looked at the demographics of the users ~\cite{yang12,li-etal-2019-rumor} or interaction periodicity \cite{kwon13a} to determine the credibility of users. Yang et al.~\shortcite{yang12} collected the user characteristics who were engaged in spreading fake news and used only user characteristics to build a fake news detector by classifying the propagation path. Li et al.~\shortcite{li-etal-2019-rumor} used a combination of user information and content features to train a LSTM with multi-task learning objective.

In this paper, we work on detecting rumor and fake news solely from the post and comments. Unlike~\citet{li-etal-2019-rumor,yang12}, we do not use the identities of user accounts. Most closely related to our work is Ma et al.~\shortcite{ma16,ma17,ma18} and Kumar et al. ~\shortcite{kumar19}. Ma et al.~\shortcite{ma16,ma17,ma18} used recurrent neural network, propagation tree kernels and recursive neural trees to model posts in the same thread either sequentially or with a tree structure to predict whether the sequence pertains to rumour or real news. Kumar et al.~\shortcite{kumar19} used various variants of LSTM (Branch LSTM, Tree LSTM and Binarized
Constituency Tree LSTM). In this paper, instead of recursive tree models, we propose a transformer network for rumor detection.

\section{Rumor Detection} \label{motivation_and_analysis}

In this section, we first define our problem statement. We address the problem of predicting the veracity of a claim given all of its responding tweets and the interactions between these tweets. We define each thread to be:
\begin{eqnarray*}
X = \{x_1, x_{2}, x_{3}, ... , x_{n}\},
\end{eqnarray*}
where $x_1$ is the source tweet, $x_{i}$ the $i^{th}$ tweet in chronological order, and $n$ the number of tweets in the thread. 

Besides the textual information, there is also structural information in the conversation tree which could be exploited \cite{ma18,wu15a}. In the tree-structured models, a pair of tweets $x_{i}$ and $x_{j}$ are only related if either $x_{i}$ is replying to $x_{j}$ or vice versa. In all of our proposed models, we allow any post to attend to any other post in the same thread. In our structure aware models, we label the relations between any pair of tweets $x_{i}$ and $x_{j}$ with a relation label, $R(i,j) \in$ \{parent, child, before, after, self\}. The value of $R(i,j)$ is obtained by applying the following set of rules in sequence: 
\textbf{parent}: if $x_{i}$ is directly replying $x_{j}$, \textbf{child}: if $x_{j}$ is directly replying $x_{i}$, \textbf{before}: if $x_{i}$ comes before $x_{j}$, \textbf{after}: if $x_{i}$ comes after $x_{j}$, \textbf{self}: if $i=j$.

The rumor detection task reduces to learning to predict each $(X,R)$ to its rumor class $y$. We conducted experiments on two rumor detection data sets, namely the Twitter15 and Twitter16 data, and the PHEME 5 data. The classes are different for the data sets we are working on:
\begin{itemize*}
\item for Twitter15 and Twitter16: $y\in$ \{\nonrumor{}, \falserumor{}, \truerumor{},  \unverified{}\}, and 
\item for PHEME, $y \in $ \{\falserumor{}, \truerumor{},  \unverified{}\}. 
\end{itemize*}

\section{Approaches} \label{rumor_detection_transformer}

For completeness, we first provide a brief description of recursive neural networks and the attention mechanism in transformer networks. We then describe the proposed models that is primarily based on the attention mechanism. 

\subsection{Recursive Neural Networks}
\label{rvnn}

Ma et al.~\shortcite{ma18} applied tree-structured recursive neural networks (RvNN) to the rumor detection problem: each thread is represented as a tree where the root is the claim, and each comment is a child to the post it is responding to.
In RvNN,  the input  of  each  node  in  the  tree is a vector of tf-idf values of the words in the post. Ma et al.~\shortcite{ma18} studied two models, a bottom-up and a top-down tree models. Kumar et al.~\shortcite{kumar19} applied a similar mechanism.The formulation is appealing as we expect information from denying and questioning comments to be propagated and used for the classification of rumors. 

However, as we will see in the data statistics in Table~\ref{tab:data_tree_analysis}, the trees in the data sets are very shallow, with the majority of comments replying directly to the source tweet instead of replying to other tweets. We found that in social media, as often the entire thread is visible, a user replying to the root post might be continuing the conversation of earlier users and not specifically writing a reply to the root post. Tree models that do not explicitly model every possible pairwise interaction between tweets are therefore sub-optimal in modelling social media conversations. Therefore, we propose to overcome the limitations with our described models in subsequent sections. 

\subsection{Transformer Networks}
The attention mechanism in transformer networks enable effective modeling of long-range dependencies~\cite{DBLP:journals/corr/VaswaniSPUJGKP17}.
This is an advantage for rumor detection since there could be many replying tweets in a conversation and it is vital to be able to model the interactions between the tweets effectively. We briefly discuss the multi-head attention (MHA) layer in the transformer network and refer the reader to~\cite{DBLP:journals/corr/VaswaniSPUJGKP17} for more details. 
Each MHA layer in the transformer network is made up of a self-attention sublayer and fully connected feed-forward sublayer.
In the self-attention sublayer, a query and a set of key-value pair is mapped to an output. The output is a weighted sum of the values, where the weight for each value is determined by the compatibility between the query and the corresponding key. The compatibility, $\alpha_{ij}$, is the attention weight between a query from position $i$ and a key from position $j$, and is calculated with simple scaled dot-product attention: 

\begin{equation}
\alpha_{ij} = \mathrm{Compatibility}(q_{i}, k_{j}) = \mathrm{softmax}(\frac{q_{i}k_{j}^T}{\sqrt{d_{k}}}) \label{compatibility_function}
\end{equation}

The output at each position, $z_i$, would then be a weighted sum of the compatibility and value at other positions. 
\begin{equation}
z_i = \sum_{j=1}^{n} \alpha_{ij} v_j, \label{weighted_sum}
\end{equation}
where $\alpha_{ij} \in [0,1]$ is the attention weight from Equation~\ref{compatibility_function}. A higher value indicates higher compatibility. 

In order to allow the model to jointly attend to information from different representation subspaces at different positions, Vaswani et al.~\shortcite{DBLP:journals/corr/VaswaniSPUJGKP17} introduced the concept of multi-head attention. Queries, keys and values are projected $h$ times with different learned linear projections. Each projected versions of queries, keys and values would perform the attention function (Equation ~\ref{weighted_sum}) in parallel, yielding $h$ output values. These are concatenated and once again projected, generating the final values. The final layers would then be passed through a fully connected feed-forward sublayer consisting of two linear layers with a RELU activation between them. 

\subsection{Post-Level Attention Network (PLAN)} 
\label{sec:plan}

\begin{figure}
\centering
\begin{subfigure}{0.4\columnwidth}
\centering
\includegraphics[width=\columnwidth]{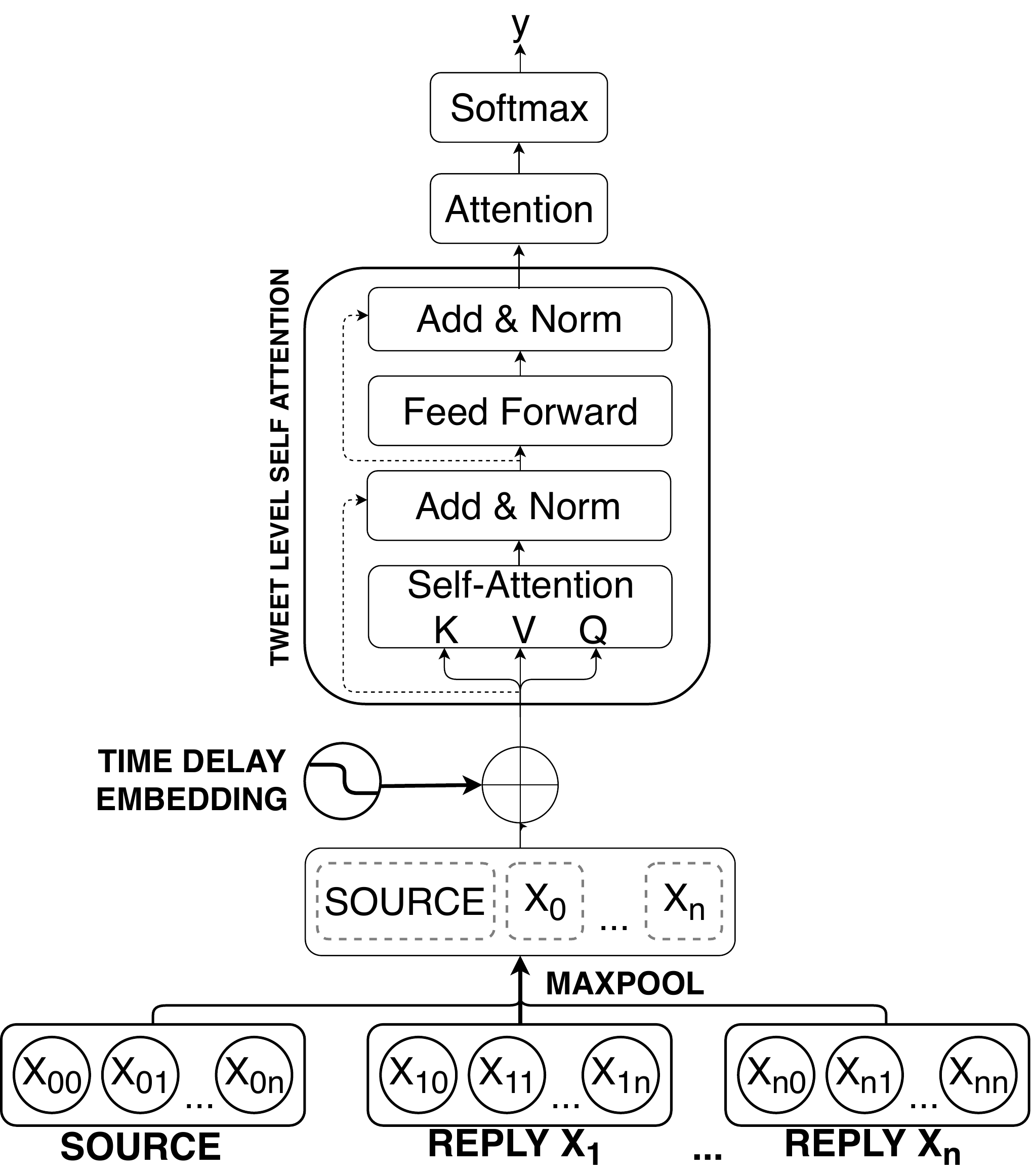}
\caption{PLAN Model}
\label{fig:plan}
\end{subfigure}
~
\begin{subfigure}{0.5\columnwidth}
\centering
\includegraphics[width=\columnwidth]{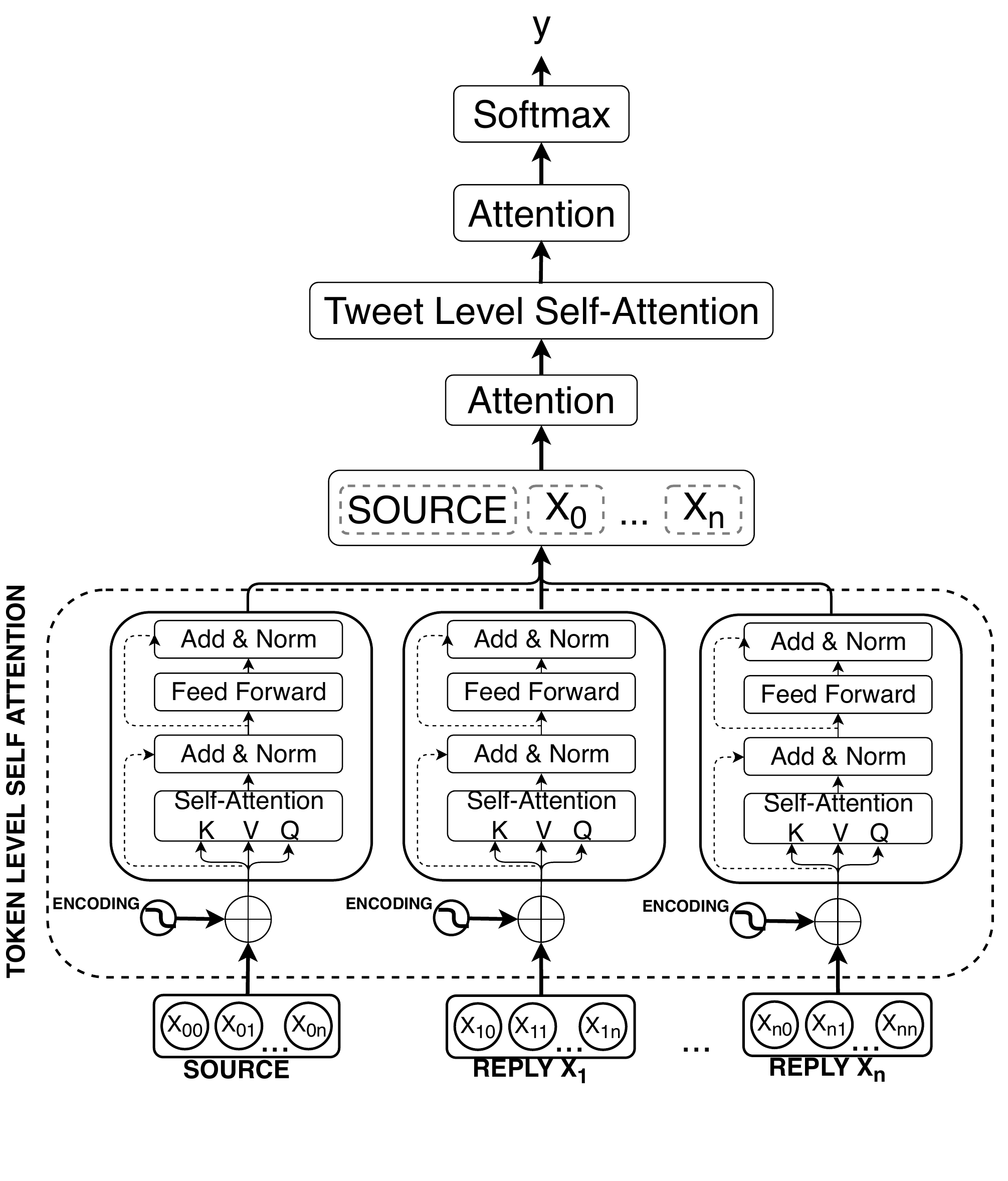}
\caption{StA-HiTPLAN Model}
\label{fig:hierachical}
\end{subfigure}
\caption{Proposed models} 
\end{figure}

The architecture of our post-level attention network (PLAN) is shown in Figure~\ref{fig:plan}. We flattened the structure of the conversation tree and arranged the tweets chronologically in a linear structure with the source tweet as the first tweet. For our PLAN model, we applied max pooling to each tweet $x_i$ in the linear structure to obtain it's sentence representation $x'_i$ . We then pass a sequence of sentence embedding $X' = (x'_{1} , x'_{2} , ..., x'_{n})$ through $s$ number of multi-head attention (MHA) layers to model the interactions between the tweets. We refer to these MHA layers as post-level attention layers. As such, this transforms $X' = (x'_{1}, x'_{2}, ..., x'_{n})$ to $U = (u_{1}, u_{2} , ..., u_{n})$ .
Lastly, we used the attention mechanism to interpolate the tweets before passing through a fully connected layer for prediction.
\begin{eqnarray}
\alpha_{k} = \mathrm{softmax}(\gamma^{T}u_{k}), \\
v = \sum_{k = 0}^{m} \alpha_{k}u_{k}, \\
p = \mathrm{softmax}(W_{p}^{T}v + b_{p}),
\end{eqnarray}
where $\gamma \in\mathbb{R}^{d_{model}}$, $\alpha_{k} \in\mathbb{R}$, $W_{p} \in\mathbb{R}^{d_{model}, K}$, $b \in\mathbb{R}^{d_{model}}$, $u_{k}$ is the output after passing through $s$ number of MHA layers, $K$ is the number of output classes, $v$ and $p$ are the representation vector and prediction vector for $X$ respectively.

\subsection{Structure Aware Post-Level Attention Network (StA-PLAN)}
\label{sec:staplan}

One possible limitation of our model is that we lose structural information by organising tweets in a linear structure. Structural information that are inherently present in a conversation tree might still be useful for fake news detection. Tree models are superior in this aspect since the structural information is modelled explicitly. To combine the strengths of tree models and the self-attention mechanism, we extended our PLAN model to include structural information explicitly. We adopted the formula in \citet{shaw-etal-2018-self} to perform structure aware self-attention:

\begin{eqnarray}
\alpha_{ij} = \mathrm{softmax}(\frac{q_{i}k_{j}^T + a_{ij}^K}{\sqrt{d_{k}}}) \label{compatibility_function_with_structure}, \\
z_i = \sum_{j=1}^{n} \alpha_{ij} (v_j + a_{ij}^V) \label{weighted_sum_with_structure}
\end{eqnarray} 
Equation \ref{compatibility_function_with_structure} extends upon Equation~\ref{compatibility_function} by explicitly adding $a_{ij}^K$ when computing compatibility. Likewise, Equation \ref{weighted_sum_with_structure} extends upon Equation~\ref{weighted_sum} by adding $a_{ij}^V$ when computing the output vector.
Both $a_{ij}^V$ and $a_{ij}^K$ are vectors that represents one of the five possible structural relationship between the pair of the tweets (i.e. parent, child, before, after and self) as described in Section~\ref{motivation_and_analysis}. These vectors are learned parameters in our model. We learn two distinct vectors to ensure the suitability for use of these vectors in the two different equations.
Intuitively, $a_{ij}^K$ gives the compatibility function more information to better determine compatibility; compatibility is now determined by both textual content and structural relationship of a pair of tweet instead of solely textual content. The addition of $a_{ij}^V$ allows both structural and content information to be propagated to other tweets. 

\subsection{Structure Aware Hierarchical Token and Post-Level Attention Network (StA-HiTPLAN)}
\label{sec:stahitplan}
Our PLAN model performs max-pooling to get the sentence representation of each tweet. However, it could be more ideal to allow the model to learn the importance of the word vectors instead. Hence, we propose a hierarchical attention model - attention at a token-level then at a post-level. An overview of the hierarchical model is shown in Figure~\ref{fig:hierachical}.

Instead of using max-pooling to obtain sentence representation, we performed token-level self-attention before using the attention mechanism to interpolate the output. This approach would also learn a more complex sentence representation for each tweet. More formally, each tweet could be represented as a sequence of word tokens where  $x_{i} = (x_{i,1}, x_{i,2}, ..., x_{i,|x_i|})$. We passed the sequence of word tokens in a tweet through $s_{word}$ number of MHA layers. This allows interactions between tokens in a tweets and we refer to these layers as token-level attention layers. After which, we used the attention mechanism to interpolate the output from the MHA layers to obtain a sentence representation for each tweet. The sentence embedding for each tweet will then be used to perform structure aware post-level self-attention as described in Section~\ref{sec:staplan}.

\subsection{Time Delay Embedding}
\label{sec:timedelay}
Tweets created at different time intervals could be interpreted differently. Tweets expressing disbelief when a source claim is first created could be common as the claim may have not been verified. However, doubtful tweets at later stage of propagation could indicate a high tendency that the source claim is fake. Therefore, we investigated the usefulness of encoding tweets with time delay information with all three of our proposed models - PLAN, StA-PLAN and StA-HiTPLAN.

To include time delay information for each tweet, we bin the tweets based on their latency from the time the source tweet was created. We set the total number of time bins to be 100 and each bin represents a 10 minutes interval. Tweets with latency of more than 1,000 minutes would fall into the last time bin.
We used the positional encoding formula introduced in the transformer network in \citet{DBLP:journals/corr/VaswaniSPUJGKP17} to encode each time bin. The time delay embedding would be added to the sentence embedding of tweet. The time delay embedding, $\mathrm{TDE}$, for each tweet is: 
\begin{eqnarray}
\mathrm{TDE}_{pos,2i} &=& \sin \frac{pos}{10000^{2i/d_{model}}}, \\ 
\mathrm{TDE}_{pos,2i+1} &=& \cos \frac{pos}{10000^{2i/d_{model}}},
\end{eqnarray}
where $pos$ represents the time bin each tweet fall into and $pos \in [0,100)$, $i$ refers to the dimension and $d_{model}$ refers to the total number of dimensions of the model. 
\newcommand{\tabitem}{~~\llap{\textbullet}~~}
\section{Experiments and Results}

We evaluate our model based on two rumour detection data sets - (i) Twitter15 and Twitter16 and (ii) PHEME 5 events data set. We show the statistics of the data sets in Table~\ref{tab:data_tree_analysis}. 

\subsection{Data and Pre-processing} \label{data_pre_processing}

\begin{table}
\centering
\small
\begin{tabular}{lccc} \toprule
Data set    & Twitter15 & Twitter16 & PHEME \\ \midrule
Tree-depth  & 2.80      & 2.77      & 3.12  \\
Num leaves  & 34.2      & 31.9      & 10.3  \\
Num tweets  & 40.6      & 37.9      & 14.9  \\ 
False       & 334       & 172       & 393 \\
True        & 350       & 189       & 1008 \\
Unverified  & 358       & 190       & 571 \\
Non-rumor   & 371       & 205       & - \\
Total trees & 1413      & 756       & 1972 \\
Total tweets& 57,368    & 27,652    & 29,383\\
\bottomrule
\end{tabular}
\caption{Average tree depths, number of leaves and tweets.}
\label{tab:data_tree_analysis}
\end{table}

For the PHEME 5 data set, we follow the experimental setting of~\citet{kumar19} in using event wise cross-validation.
For the Twitter15 and Twitter 16 data sets, there is a large proportion of retweets in each claim: 89\% for Twitter15 and 90\% for Twitter16. As we assume that retweets do not contribute new information to the model, we removed all retweets for Twitter15 and Twitter16. After the removal of retweets, we observe that a small number of claims would be left with only the source tweet. Since the principle behind our methodology is that we could exploit crowd signals for rumor detection, claims without any replies should then be ``unverified''. Hence, we amended the labels for such claims to be ``unverified'' in the training data (1.49\% of Twitter15 - 8 False, 10 True and 3 Non-Rumour. 1.46\% of Twitter16 - 9 False, 2 True and 0 Non-Rumour). In order for our results to be comparable with previous work, we excluded such claims from our testing set. We used the original splits released by~\cite{ma18} to split our data.  
We show the statistics of the data sets after pre-processing in Table~\ref{tab:data_tree_analysis}.

\subsection{Experimental Setup}

\begin{table*}[ht]
\small
\centering
\begin{tabular}{ccccccccccc} \toprule
& \multicolumn{5}{c}{Twitter15} & \multicolumn{5}{c}{Twitter16} \\
Method		&	Accuracy & F & T & U & NR & Accuracy & F & T & U & NR \\ \midrule
BU-RvNN	(Original) & 70.8 & 72.8 & 75.9 & 65.3 & 69.5
& 71.8 & 71.2 & 77.9 & 65.9 & 72.3\\
TD-RvNN (Original) & 72.3 & 75.8 & 82.1 & 65.4 & 68.2 
& 73.7 & 74.3 & 83.5 & 70.8 & 66.2 \\
BU-RvNN (Ours)	& 70.5 & 71.0 & 72.1 & 73.0 & 65.5 
& 80.6 & 75.5 & 89.3 & 83.0 & 73.4\\
TD-RvNN	(Ours) & 65.9 & 66.1 & 68.9 & 71.4 & 55.9 
& 76.7 & 69.8 & 87.2 & 81.3 & 66.1 \\
\midrule
\textbf{PLAN}	& 84.5 & \textbf{85.8} & \textbf{89.5} & 80.2 & 82.3 
 & \textbf{87.4} & \textbf{83.9} & 91.7 & \textbf{88.8} & \textbf{85.3} \\
\textbf{StA-PLAN} & \textbf{85.2} & 84.6 & 88.4 & \textbf{83.7} & 84.0
& 86.8 & 83.3 & \textbf{92.7} & \textbf{88.8} & 82.6\\
\textbf{StA-HiTPLAN} & 80.8 & 80.2 & 85.1 & 76.0 & 81.7 
& 80.7 & 76.5 & 88.8 & 82.0 & 74.9 \\
\textbf{PLAN} \textbf{+ time-delay}	& 84.1 & 84.2 & 87.3 & 80.3 & 84.2
& 84.8 & 77.6 & 89.7 & 85.6 & 84.9\\
\textbf{StA-PLAN} \textbf{+ time-delay} & 85.0 & 85.7 & 88.3 & 81.4 & \textbf{84.4}
 & 86.6 & 83.3 & 92.3 & 86.6 & 84.2 \\
\bottomrule 
\end{tabular}
\caption{Accuracy on Twitter15 and Twitter16, where F, T, U and NR stands for False, True, Unverified and Non-rumor respectively. We report the F1-Score for each individual class. The results of rows with (Original) were referenced from~\cite{ma17}, while the remaining rows are based on our own implementation of the models.} 
\label{tab:results_15}
\end{table*}

\begin{table}
\small
\centering
\resizebox{0.9\columnwidth}!{
\begin{tabular}{cccccc} \toprule
Method & Macro F-Score \\ \midrule
Branch LSTM - Multitask	& 35.9 \\
Tree LSTM - Multitask & 37.9 \\
BCTree LSTM - Multitask & 37.1 \\ \bottomrule
\textbf{PLAN}	& 36.0 \\
\textbf{StA-PLAN} & 34.9 \\
\textbf{StA-HiTPLAN} & 37.9 \\
\textbf{PLAN + Time Delay}	& 38.6 \\
\textbf{StA-PLAN + Time Delay} & 36.9 \\
\textbf{StA-HiTPLAN + Time Delay} & \textbf{39.5} \\
\textbf{StA-HiTPLAN + Time Delay (Random split)} & 77.4 \\
\bottomrule 
\end{tabular}
}
\caption{F1-score on PHEME. We used the same train-test splits as ~\cite{kumar19} (Except the last row) and the results of the first three rows were referenced from the paper.}
\label{tab:results_pheme}
\end{table}

In all experiments, we used the GLOVE 300d \cite{pennington2014glove} embedding to represent each token in a tweet. (Preliminary experiments with BERT~\cite{devlin2018bert} did not improve results, and were computationally far more expensive than GLOVE). We used the same set of hyper parameters in all of our experiments for both PHEME and Twitter15 and Twitter16. Our model dimension is 300 and the dimension of intermediate output is 600. We used 12 post-level MHA layers and 2 token-level MHA layers. For training of the model, we used the ADAM optimizer with 6000 warm start-up steps. We used an initial learning rate of 0.01 with 0.3 dropout. We used a batch size of 32 for PLAN and StA-PLAN, and 16 for StA-HiTPLAN due to memory limitations of the GPUs used. We compare the following models:
\begin{itemize*}
\item PLAN: The post-level attention network in Section~\ref{sec:plan}.
\item StA-PLAN: The structure aware post-level attention network in Section~\ref{sec:staplan}.
\item StA-HiTPLAN: The structure aware hierarchical attention model in Section~\ref{sec:stahitplan}.
\end{itemize*}
We also investigate these models with and without the time-delay embedding as described in Section~\ref{sec:timedelay}. However, as time delay information did not improve upon PLAN and StA-PLAN for Twitter15 and Twitter16, we did not run experiments for StA-HiTPLAN + time delay for Twitter15 and Twitter16. 
For Twitter15 and Twitter 16, we compared our proposed models with RvNN models proposed by Ma et al.~\shortcite{ma18}. As we are using different preprocessing steps (described in ~\ref{data_pre_processing}) for Twitter15 and Twitter16, we retrained the RvNN models with our implementation. We report both the original results and results from our implementation. For PHEME, we compare against the LSTM models proposed by Kumar et al.~\shortcite{kumar19}. As there were several combination of embeddings and models in~\cite{kumar19}, we compare our results with the best results reported for each variant of LSTM model proposed.

We summarize our experimental results in Tables~\ref{tab:results_15} and \ref{tab:results_pheme}. 
For Twitter 15 and Twitter 16, all our models outperform tree recursive neural networks, and our best models outperform by 14.2\% for Twitter15 and 6.8\% for Twitter16. For PHEME, our best model outperforms previous work by 1.6\% F1-score. 

\subsection{Discussion of results} \label{Results Analysis}
In the rest of this section, we analyze the performance of our proposed models on Twitter15, Twitter16 and PHEME. We found conflicting conclusions for the different datasets. In particular, although all of the datasets are similar in nature, the state-of-the-art performance of PHEME is far worse than that of Twitter15 and Twitter16. As such, we also provide analysis suggesting possible explanations for this disparity in results for the two datasets.

\subsubsection{Structural Information:} 
We proposed two methods - StA-PLAN and time-delay embeddings that aim to capture structural and chronological information among posts. StA-PLAN is the best performer for Twitter15 but did not outperform PLAN for Twitter16, though results were not substantially different. The reason structural aware model only works for Twitter15 might be because the Twitter15 data set is substantially bigger (in terms of number of tweets, see Table~\ref{tab:data_tree_analysis}) than both PHEME and Twitter16. A big data set might be necessary to exploit the complicated structural information in the structure aware model. Time-delay information was useful for PHEME, where all proposed models performed better with time delay. On the other hand, time delay information was not useful for Twitter15 and Twitter16. Overall, it is unclear if structural information is useful for these data sets, and we leave further investigation to future work.

\begin{table}
\centering
\small
\begin{tabular}{cccc} \toprule
      Words    & Twitter15 & Twitter16 & PHEME \\
 \midrule
 ``{\it true}'' & 20.4	& 17.6	& 9.5 \\
  ``{\it real}'' & 34.4	& 25.4	& 11.7 \\
  ``{\it fake}'' & 14.1	& 10.9	& 2.2    \\
  ``{\it lies}'' & 5.9	& 17.6	& 2.1\\     
\bottomrule
\end{tabular}
\caption{Percentage of claims containing each word.}
\label{tab:word_analysis}
\end{table}

\subsubsection{Token level self-attention:} We proposed a token level self-attention mechanism to model the relationship between tokens in a tweet with our StA-HiTPLAN model. StA-HiTPLAN was the best performer for PHEME but did not outperform our baseline model for Twitter15 and Twitter16. 

To investigate why StA-HiTPLAN was the best performer for PHEME, we hypothesize that the signals for rumor detection in PHEME could be much weaker or expressed in a more implicit manner. Therefore, it would be necessary to study the interaction between the words to better capture the meaning of the whole sentence. To this end, we identified individual words that could be useful in determining veracity of the claim. We compute the statistics of the words ``{\it true}'', ``{\it real}'' , ``{\it fake}'' and ``{\it lies}'' in the three data sets as shown in Table~\ref{tab:word_analysis}. These words act as a proxy for crowd signals for the model to learn from and we observe that the usage of such words is the lowest in PHEME. It was also pointed out in~\cite{kumar19} that most of the replying tweets in PHEME were largely neutral comments. Therefore, these observations suggests that there is weaker crowd signal in PHEME. Hence, token level attention might have been necessary to do well in the PHEME. 
We provide an example where token-level attention is useful for PHEME in Figure~\ref{fig:heatmap}. The important tweet shown in the example is however not straightforward and would require inferring that "doesn't need facts to write stories" would imply that the claim is fake. Therefore, token-level self-attention is required to accurately capture the meaning of the phrase.

We further analyze the performance of our model on the PHEME dataset in the section below. 

\begin{table*}[ht]
\centering
\small
\resizebox{1.8\columnwidth}!{
\begin{tabular}{p{5cm}lp{9.4cm}cp{0.4cm}}  \toprule
\textbf{(Label) Claim} & \multicolumn{2}{l}{\textbf{Important Tweets}} & \textbf{\#Tweets}  \\ \midrule 
\multirow{5}{4cm}{
(\textsc{Unverified}) Surprising number of vegetarians secretly eat meat}
& 1 &
@HuffingtonPost ........ then they aren't vegetarians.  & 33 \\
& 2 &
@HuffingtonPost this article is stupid. If they ever eat meat, they are not vegetarian.\\
& 3 &
@HuffingtonPost @laurenisaslayer LOL this could be a The Onion article \\ \hline
\multirow{5}{4cm}{
(\textsc{True}) Officials took away this Halloween decoration after reports of it being a real suicide victim. It is still unknown. URL}
& 1 & 
@NotExplained how can it be unknown if the officials took it down...... They have to touch it and examine it & 46 \\
& 2 & 
@NotExplained did anyone try walking up to it to see if it was real or fake? this one seems like an easy case to solve \\
& 3 &
@NotExplained thats from neighbours \\ \hline
\multirow{5}{5cm}{
(\textsc{False}) CTV News confirms that Canadian authorities have provided US authorities with the name Michael Zehaf-Bibeau in connection to Ottawa shooting}
& 1 & 
@inky\_mark @CP24 as part of a co-op criminal investigation one would URL doesn't need facts to write stories it appears. & 5 \\
& 2 & 
@CP24 I think that soldiers should be armed and wear protective vests when they are on guard any where. \\
& 3 & 
@CP24 That name should not be mentioned again. \\
\bottomrule
\end{tabular}
}
\caption{Samples of tweet level explanation for each claim. We sort tweets based on the number of times it was identified as the most relevant tweet to the most important tweet and show the top three tweets. The right most column gives the number of tweets in the thread.}
\label{fig:tweet_level_anaysis}
\end{table*}

\subsection{Analyzing results for PHEME}
\label{PHEME analysis} 
In this section, we provide some error analysis and intuition on the disparity between the performance for Twitter15, Twitter16 and PHEME.

\subsubsection{Out-of-domain classification}
\label{out-of-domain classification}
One key difference between Twitter15 and Twitter16 and PHEME is that the train-test split for PHEME was done at an event level whereas Twitter15 and Twitter16 were split randomly. As there are no overlapping events in the training and testing data for PHEME, the model essentially has to learn to perform cross-domain classification. As the model would have naturally learnt event specific features from the train set, these event specific features would result in poorer performance on test set from another event. To verify our hypothesis, we trained and tested the StA-HiTPLAN + time-delay model by splitting the train and test set randomly. As seen in Table \ref{tab:results_pheme}, we were able to obtain an F-score of 77.4. (37.5 higher compared to using events split). Because splitting by events a more realistic setting, we would explore methods to make our model event agnostic in future works.

\subsection{Explaining the predictions} \label{explainability}
One key advantage of our model is that we could examine the attention weights in our model to interpret the results of our model. We illustrate how we can generate both token-level and tweet-level explanations for our predictions.

\subsubsection{Post-Level Explanations}
\label{post_level_explanations}
As described in Section~\ref{rumor_detection_transformer}, we used the self-attention mechanism to aggregate information among tweets. The amount of information a tweet is propagating to another tweet is weighted by the relatedness between the pair, measured by the self-attention weight between them - higher self-attention weight imply higher relatedness. We then use the attention mechanism to interpolate tweets in the final layer before the prediction layer. This generates a representation vector for the claim that would be used for prediction. The attention weight for each tweet indicates the level of importance the model has placed on that tweet for prediction. Higher attention weight implies higher importance. 
Therefore, to generate the explanations for each claim, we obtain the tweet with the highest attention weight at the final layer - this would give us the most important tweet, $tweet_{impt}$, for prediction. After which, we obtained the most relevant tweet, $tweet_{rel,i}$ to $tweet_{impt}$ at the $i^{th}$ MHA layer. We do so by obtaining tweets with highest self-attention weight with this particular tweet at each MHA layer. The same tweet could be identified as the most relevant tweet multiple times. We ranked each tweet based on the number of times it was identified as a relevant tweet. The top three tweets would be the explanation for this particular claim. Table \ref{fig:tweet_level_anaysis} shows examples of the top three tweets: we see replies that cast doubt on, or confirm, the claim. These replies were accurately identified and used by our model in debunking or confirming the factuality of claims despite a large number of other tweets present in the conversation.

\subsubsection{Token-Level Explanation}
\begin{figure}[t]
\centering
\includegraphics[width= 0.8 \columnwidth]{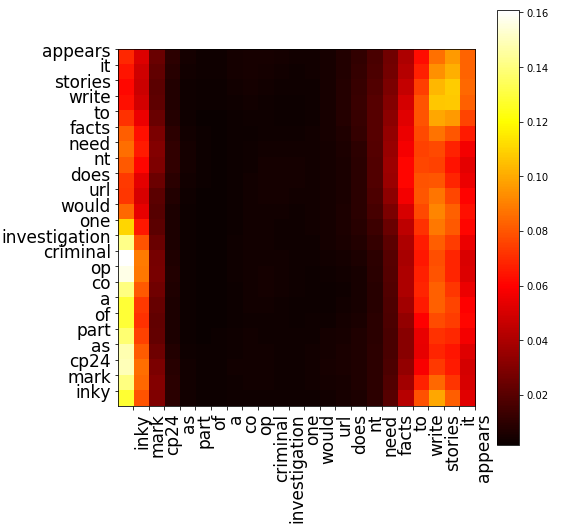}
\caption{Heatmap showing the important tokens with token-level self-attention for a fake claim in PHEME. A lighter colour means higher importance.}
\label{fig:heatmap}
\end{figure}

As described in Section~\ref{post_level_explanations}, we could extract the important tweets to a prediction - These tweets provide tweet-level explanations for our model. Following the steps described, we obtained \textit{"@inky\_mark @CP24 as part of a co-op criminal investigation one would URL doesn't need facts to write stories it appears."} as the most important tweet to predict the fake claim, \textit{"CTV News confirms that Canadian authorities have provided US authorities with the name Michael Zehaf-Bibeau in connection to Ottawa shooting"} correctly. This claim was obtained from the PHEME dataset. As shown in Figure~\ref{fig:heatmap}, most tokens in the tweet have equally high attention weights with tokens at the end of the tweet. A further examination of the attention weights show that high weights were placed on the phrase "facts to write stories it appears". As such, "facts to write stories it appears" could be deemed as an important phrase that could explain the prediction of our model for this claim.

\section{Conclusion}

We have proposed three models that outperformed state-of-the-art models on three data sets. Our model utilizes the self-attention mechanism to model pairwise interactions between posts. We utilized the attention mechanism to provide possible explains of the prediction by extracting the important posts that resulted in the prediction. We also investigated mechanisms to capture structure and time information.

In this paper, we focused only on data with community response. Recent papers that perform rumor detection with user identity information have shown superior results.
Lastly, another direction in fake news detection is fact checking with reliable sources. Fact checking and rumor detection could provide complementary information and could be done in a joint manner. We would consider this in future.

\section*{Acknowledgement}
We would like to thank all the anonymous reviewers for their help and insightful comments. This piece of work was done when Qian Zhong was at SMU.

\fontsize{9.0pt}{10.0pt} \selectfont

\bibliography{rumor}

\begin{thebibliography}{}

\bibitem[\protect\citeauthoryear{Castillo, Mendoza, and
  Poblete}{2011}]{castillo11}
Castillo, C.; Mendoza, M.; and Poblete, B.
\newblock 2011.
\newblock Information credibility on twitter.
\newblock In {\em Proceedings of the 20th International Conference on World
  Wide Web}.

\bibitem[\protect\citeauthoryear{Chen \bgroup et al\mbox.\egroup
  }{2017}]{DBLP:journals/corr/ChenWLZYW17}
Chen, T.; Wu, L.; Li, X.; Zhang, J.; Yin, H.; and Wang, Y.
\newblock 2017.
\newblock Call attention to rumors: Deep attention based recurrent neural
  networks for early rumor detection.
\newblock {\em CoRR} abs/1704.05973.

\bibitem[\protect\citeauthoryear{Ciampaglia \bgroup et al\mbox.\egroup
  }{2015}]{giovanni15}
Ciampaglia, G.~L.; Shiralkar, P.; Rocha, L.~M.; Bollen, J.; Menczer, F.; and
  Flammini, A.
\newblock 2015.
\newblock Computational fact checking from knowledge networks.
\newblock {\em CoRR} abs/1501.03471.

\bibitem[\protect\citeauthoryear{Devlin \bgroup et al\mbox.\egroup
  }{2018}]{devlin2018bert}
Devlin, J.; Chang, M.-W.; Lee, K.; and Toutanova, K.
\newblock 2018.
\newblock Bert: Pre-training of deep bidirectional transformers for language
  understanding.
\newblock {\em arXiv preprint arXiv:1810.04805}.

\bibitem[\protect\citeauthoryear{Enayet and El-Beltagy}{2017}]{omar17}
Enayet, O., and El-Beltagy, S.~R.
\newblock 2017.
\newblock Niletmrg at semeval-2017 task 8: Determining rumour and veracity
  support for rumours on twitter.
\newblock In {\em Proceedings of the 11th International Workshop on Semantic
  Evaluation (SemEval-2017)}.

\bibitem[\protect\citeauthoryear{Fuller, Biros, and Wilson}{2009}]{fuller09}
Fuller, C.~M.; Biros, D.~P.; and Wilson, R.~L.
\newblock 2009.
\newblock Decision support for determining veracity via linguistic-based cues.
\newblock {\em Decis. Support Syst.} 46(3).

\bibitem[\protect\citeauthoryear{Funke}{2019}]{poynter_2019}
Funke, D.
\newblock 2019.
\newblock Snopes pulls out of its fact-checking partnership with facebook.

\bibitem[\protect\citeauthoryear{Gunther, Beck, and
  Nisbet}{2018}]{gunther2018fake}
Gunther, R.; Beck, P.~A.; and Nisbet, E.~C.
\newblock 2018.
\newblock Fake news did have a significant impact on the vote in the 2016
  election: Original full-length version with methodological appendix.
\newblock {\em Columbus, OH: Ohio State University}.

\bibitem[\protect\citeauthoryear{Hu \bgroup et al\mbox.\egroup }{2013}]{hu13}
Hu, X.; Tang, J.; Zhang, Y.; and Liu, H.
\newblock 2013.
\newblock Social spammer detection in microblogging.
\newblock In {\em Proceedings of the Twenty-Third International Joint
  Conference on Artificial Intelligence}.

\bibitem[\protect\citeauthoryear{Kochkina, Liakata, and
  Zubiaga}{2018}]{kochkina_liakata_zubiaga_2018}
Kochkina, E.; Liakata, M.; and Zubiaga, A.
\newblock 2018.
\newblock Pheme dataset for rumour detection and veracity classification.

\bibitem[\protect\citeauthoryear{Kumar and Carley}{2019}]{kumar19}
Kumar, S., and Carley, K.~M.
\newblock 2019.
\newblock Tree lstms with convolution units to predict stance and rumor
  veracity in social media conversations.
\newblock In {\em Proceedings of the Annual Meeting of the Association for
  Computational Linguistics, {ACL} 2019}.

\bibitem[\protect\citeauthoryear{Kwon \bgroup et al\mbox.\egroup
  }{2013}]{kwon13a}
Kwon, S.; Cha, M.; Jung, K.; Chen, W.; and Wang, Y.
\newblock 2013.
\newblock Prominent features of rumor propagation in online social media.
\newblock In {\em {ICDM}},  1103--1108.
\newblock {IEEE} Computer Society.

\bibitem[\protect\citeauthoryear{Li, Zhang, and Si}{2019}]{li-etal-2019-rumor}
Li, Q.; Zhang, Q.; and Si, L.
\newblock 2019.
\newblock Rumor detection by exploiting user credibility information, attention
  and multi-task learning.
\newblock In {\em Proceedings of the 57th Annual Meeting of the Association for
  Computational Linguistics}.

\bibitem[\protect\citeauthoryear{Ma \bgroup et al\mbox.\egroup }{2016}]{ma16}
Ma, J.; Gao, W.; Mitra, P.; Kwon, S.; Jansen, B.~J.; Wong, K.-F.; and Cha, M.
\newblock 2016.
\newblock Detecting rumors from microblogs with recurrent neural networks.
\newblock In {\em Proceedings of the Twenty-Fifth International Joint
  Conference on Artificial Intelligence}, IJCAI'16.

\bibitem[\protect\citeauthoryear{Ma, Gao, and Wong}{2017}]{ma17}
Ma, J.; Gao, W.; and Wong, K.-F.
\newblock 2017.
\newblock Detect rumors in microblog posts using propagation structure via
  kernel learning.
\newblock In {\em Proceedings of the 55th Annual Meeting of the Association for
  Computational Linguistics (Volume 1: Long Papers)}.

\bibitem[\protect\citeauthoryear{Ma, Gao, and
  Wong}{2018a}]{Ma:2018:DRS:3184558.3188729}
Ma, J.; Gao, W.; and Wong, K.-F.
\newblock 2018a.
\newblock Detect rumor and stance jointly by neural multi-task learning.
\newblock In {\em Companion Proceedings of the The Web Conference 2018}.

\bibitem[\protect\citeauthoryear{Ma, Gao, and Wong}{2018b}]{ma18}
Ma, J.; Gao, W.; and Wong, K.-F.
\newblock 2018b.
\newblock Rumor detection on twitter with tree-structured recursive neural
  networks.
\newblock In {\em Proceedings of the 56th Annual Meeting of the Association for
  Computational Linguistics (Volume 1: Long Papers)}.

\bibitem[\protect\citeauthoryear{Mihalcea and Strapparava}{2009}]{mihalcea09}
Mihalcea, R., and Strapparava, C.
\newblock 2009.
\newblock The lie detector: Explorations in the automatic recognition of
  deceptive language.
\newblock In {\em Proceedings of the {ACL}-{IJCNLP} 2009 Conference Short
  Papers}.

\bibitem[\protect\citeauthoryear{O'Brien \bgroup et al\mbox.\egroup
  }{2018}]{obrien18}
O'Brien, N.; Latessa, S.; Evangelopoulos, G.; and Boix, X.
\newblock 2018.
\newblock The language of fake news: Opening the black-box of deep learning
  based detectors.
\newblock In {\em Proceedings of the Neurips 2018 Workshop on AI for social
  good}.

\bibitem[\protect\citeauthoryear{Ott \bgroup et al\mbox.\egroup }{2011}]{ott11}
Ott, M.; Choi, Y.; Cardie, C.; and Hancock, J.~T.
\newblock 2011.
\newblock Finding deceptive opinion spam by any stretch of the imagination.
\newblock In {\em Proceedings of the 49th Annual Meeting of the Association for
  Computational Linguistics: Human Language Technologies - Volume 1}, HLT '11.

\bibitem[\protect\citeauthoryear{Pennington, Socher, and
  Manning}{2014}]{pennington2014glove}
Pennington, J.; Socher, R.; and Manning, C.~D.
\newblock 2014.
\newblock Glove: Global vectors for word representation.
\newblock In {\em Empirical Methods in Natural Language Processing (EMNLP)}.

\bibitem[\protect\citeauthoryear{Rapoza}{2017}]{rapoza_2017}
Rapoza, K.
\newblock 2017.
\newblock Can 'fake news' impact the stock market?

\bibitem[\protect\citeauthoryear{Rubin \bgroup et al\mbox.\egroup
  }{2016}]{rubin16}
Rubin, V.; Conroy, N.; Chen, Y.; and Cornwell, S.
\newblock 2016.
\newblock Fake news or truth? using satirical cues to detect potentially
  misleading news.
\newblock In {\em Proceedings of the Second Workshop on Computational
  Approaches to Deception Detection}.

\bibitem[\protect\citeauthoryear{Sharma \bgroup et al\mbox.\egroup
  }{2019}]{sharma2019}
Sharma, K.; Qian, F.; Jiang, H.; Ruchansky, N.; Zhang, M.; and Liu, Y.
\newblock 2019.
\newblock Combating fake news: A survey on identification and mitigation
  techniques.
\newblock {\em ACM Transactions on Intelligent Systems and Technology (TIST)}
  10(3):21.

\bibitem[\protect\citeauthoryear{Shaw, Uszkoreit, and
  Vaswani}{2018}]{shaw-etal-2018-self}
Shaw, P.; Uszkoreit, J.; and Vaswani, A.
\newblock 2018.
\newblock Self-attention with relative position representations.
\newblock In {\em Proceedings of the 2018 Conference of the North {A}merican
  Chapter of the Association for Computational Linguistics: Human Language
  Technologies, Volume 2 (Short Papers)}.

\bibitem[\protect\citeauthoryear{Thorne \bgroup et al\mbox.\egroup
  }{2018}]{thorne18}
Thorne, J.; Vlachos, A.; Cocarascu, O.; Christodoulopoulos, C.; and Mittal, A.
\newblock 2018.
\newblock The fact extraction and verification (fever) shared task.
\newblock In {\em Proceedings of the First Workshop on Fact Extraction and
  VERification (FEVER)}.

\bibitem[\protect\citeauthoryear{Vaswani \bgroup et al\mbox.\egroup
  }{2017}]{DBLP:journals/corr/VaswaniSPUJGKP17}
Vaswani, A.; Shazeer, N.; Parmar, N.; Uszkoreit, J.; Jones, L.; Gomez, A.~N.;
  Kaiser, L.; and Polosukhin, I.
\newblock 2017.
\newblock Attention is all you need.
\newblock {\em CoRR} abs/1706.03762.

\bibitem[\protect\citeauthoryear{Wang}{2017}]{wang17liar}
Wang, W.~Y.
\newblock 2017.
\newblock "liar, liar pants on fire": {A} new benchmark dataset for fake news
  detection.
\newblock In {\em Proceedings of the 55th Annual Meeting of the Association for
  Computational Linguistics, {ACL} 2017, Vancouver, Canada, July 30 - August 4,
  Volume 2: Short Papers},  422--426.

\bibitem[\protect\citeauthoryear{Wu, Yang, and Zhu}{2015}]{wu15a}
Wu, K.; Yang, S.; and Zhu, K.~Q.
\newblock 2015.
\newblock False rumors detection on sina weibo by propagation structures.
\newblock {\em 2015 IEEE 31st International Conference on Data Engineering}.

\bibitem[\protect\citeauthoryear{Yang \bgroup et al\mbox.\egroup
  }{2012}]{yang12}
Yang, F.; Liu, Y.; Yu, X.; and Yang, M.
\newblock 2012.
\newblock Automatic detection of rumor on sina weibo.
\newblock In {\em Proceedings of the ACM SIGKDD Workshop on Mining Data
  Semantics}.

\end{thebibliography}
\bibliographystyle{aaai}

\end{document}